\def\year{2020}\relax
\documentclass[letterpaper]{article} 
\usepackage{aaai20}  
\usepackage{times}  
\usepackage{helvet} 
\usepackage{courier}  
\usepackage[hyphens]{url}  
\usepackage{graphicx} 
\usepackage{graphicx}  
\usepackage{hyperref}
\usepackage{acronym}
\usepackage{amsmath}
\usepackage{booktabs}
\usepackage{multirow}
\usepackage[inline]{enumitem}
\usepackage{array}
\usepackage{verbatim}
\usepackage{latexsym}
\usepackage{amsfonts}
\usepackage{xcolor}

\newcommand{\citet}[1]{\citeauthor{#1}~\shortcite{#1}}
\newcommand{\citep}{\cite}

\acrodef{RefNet}{Reference-aware Network}
\acrodef{BBC}{Background Based Conversation}
\acrodef{RC}{Reading Comprehension}

\title{RefNet: A Reference-aware Network for Background Based Conversation}

\author{
Chuan Meng,\textsuperscript{\rm 1}
Pengjie Ren,\textsuperscript{\rm 2}\footnotemark[1]
Zhumin Chen,\textsuperscript{\rm 1}\thanks{Corresponding author}
Christof Monz,\textsuperscript{\rm 2}
Jun Ma,\textsuperscript{\rm 1}
Maarten de Rijke\textsuperscript{\rm 2}\\
\textsuperscript{\rm 1}Shandong University, Qingdao, China,\\
\textsuperscript{\rm 2}University of Amsterdam, Amsterdam, The Netherlands\\
mengchuan@mail.sdu.edu.cn\\
\{p.ren, c.monz, derijke\}@uva.nl, \{chenzhumin, majun\}@sdu.edu.cn
}

\begin{document}
\maketitle
\begin{abstract}
Existing conversational systems tend to generate generic responses.
Recently, \acp{BBC} have been introduced to address this issue.
Here, the generated responses are grounded in some background information.
The proposed methods for \acp{BBC} are able to generate more informative responses, however, they either cannot generate natural responses or have difficulties in locating the right background information.
In this paper, we propose a \ac{RefNet} to address both issues.
Unlike existing methods that generate responses token by token, \ac{RefNet} incorporates a novel \textit{reference decoder} that provides an alternative way to learn to directly select a \textit{semantic unit} (e.g., a span containing complete semantic information) from the background.
Experimental results show that \ac{RefNet} significantly outperforms state-of-the-art methods in terms of both automatic and human evaluations, indicating that \ac{RefNet} can generate more appropriate and human-like responses.
\end{abstract}

\section{Introduction}
Dialogue systems have attracted a lot of attention recently \cite{huang2019challenges}.
Sequence-to-sequence models  \cite{sutskever2014sequence,lei2018sequicity} are an effective framework that is commonly adopted in existing studies.
However, a problem of sequence-to-sequence based methods is that they tend to generate generic and non-informative responses which provide deficient information~\cite{gao2019neural}.

\begin{figure}[t]
 \centering
 \includegraphics[width=1\columnwidth]{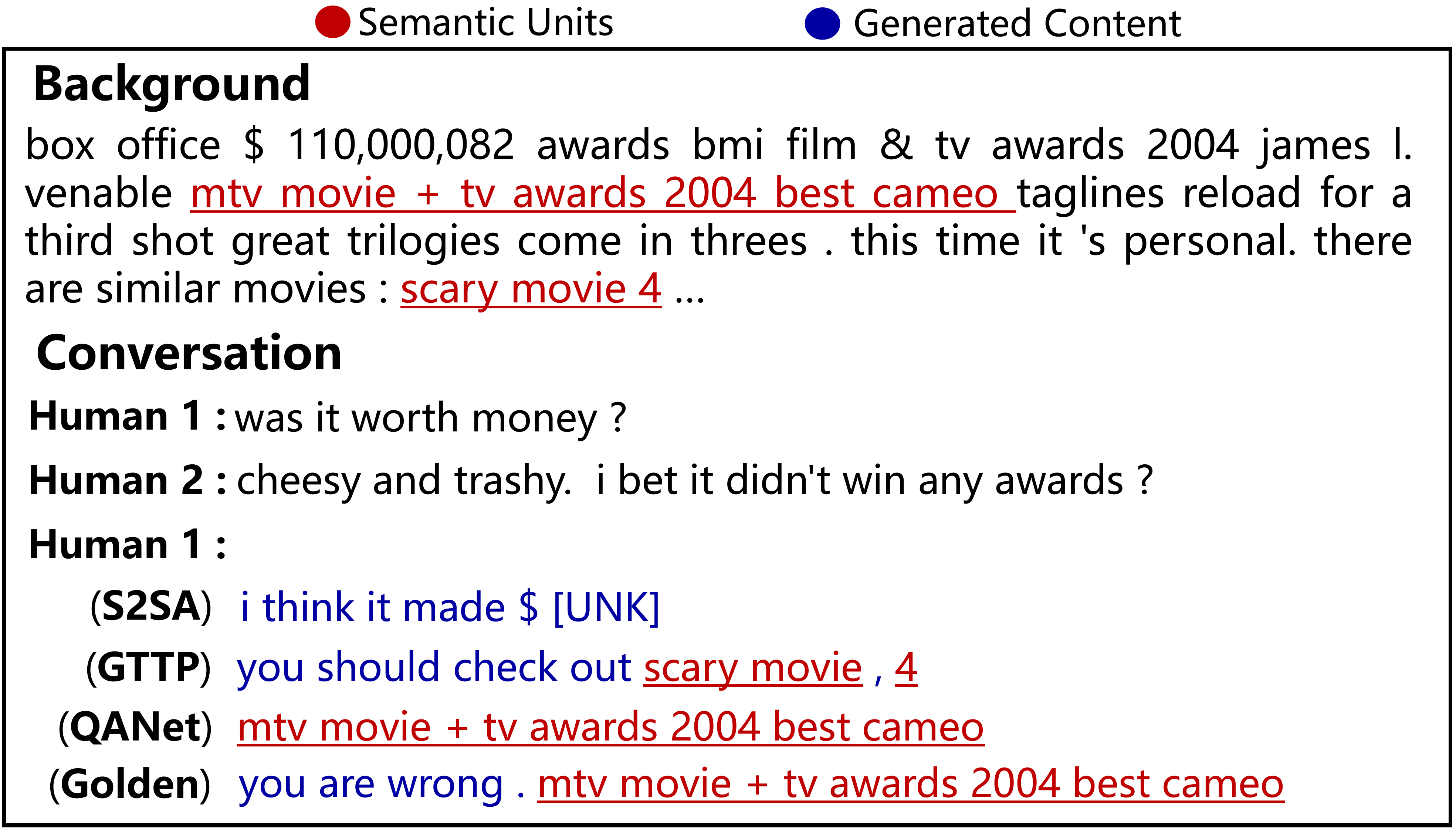}
 \caption{\acf{BBC}.}
 \label{f1}
\end{figure}

Previous research has proposed various methods to alleviate the issue, such as adjusting objective functions~\cite{li2016diversity,jiang-2019-improving}, incorporating external knowledge ~\cite{ghazvininejad2018knowledge,parthasarathi2018extending,dinan2018wizard}, etc.
Recently, \acfp{BBC} have been proposed for generating more informative responses that are grounded in some background information \cite{zhou2018dataset,moghe2018towards}.
As shown in Fig.~\ref{f1}, unlike previous conversational settings \citep{serban2016building}, in a \ac{BBC} background material (e.g., a plot or review about a movie) is supplied to promote topic-specific conversations.

Existing methods for \acp{BBC} can be grouped into two categories, \emph{generation-based} methods (e.g., GTTP \cite{see2017get}) and \emph{extraction-based} methods (e.g., QANet \cite{yu2018qanet}).
Generation-based methods generate the response token by token, so they can generate natural and fluent responses, generally.
However, generation-based methods suffer from two issues.
First, they are relatively ineffective in leveraging background information.
For example, for the case in Fig.~\ref{f1}, S2SA does not leverage background information at all.
Second, they have difficulties locating the right semantic units in the background information.
Here, a \textit{semantic unit} is a span from the background information that expresses complete semantic meaning.
For example, in Fig.~\ref{f1}, the background contains many semantic units, e.g., ``\emph{mtv movie + tvawards 2004 best cameo}'' and ``\emph{scary movie 4}.''
GTTP uses the wrong semantic unit ``\emph{scary movie 4}'' to answer the question by ``human 2.''
Moreover, because generation-based methods generate the response one token at a time, they risk breaking a complete semantic unit, e.g., ``\emph{scary movie 4}'' is split by a comma in the response of GTTP in Fig.~\ref{f1}.
The reason is that generation-based methods lack a global perspective, i.e., each decoding step only focuses on a single (current) token and does not consider the tokens to be generated in the following steps.
Extraction-based methods extract a span from the background as their response and are relatively good at locating the right semantic unit.
But because of their extractive nature, they cannot generate natural conversational responses, see, e.g., the response of QANet in Fig.~\ref{f1}.

We propose a \textbf{Ref}erence-aware \textbf{Net}work (RefNet) to address above issues.
\ac{RefNet} consists of four modules: a \emph{background encoder}, a \emph{context encoder}, a \emph{decoding switcher}, and a \emph{hybrid decoder}.
The background encoder and context encoder encode the background and conversational context into representations, respectively.
Then, at each decoding step, the decoding switcher decides between \emph{reference decoding} and \emph{generation decoding}.
Based on the decision made by the decoding switcher, the hybrid decoder either selects a semantic unit from the background (\emph{reference decoding}) or generates a token otherwise (\emph{generation decoding}).
In the latter case, the decoding switcher further determines whether the hybrid decoder should predict a token from the vocabulary or copy one from the background.
Besides generating the response token by token, \ac{RefNet} also provides an alternative way to learn to select a semantic unit from the background directly.
Experiments on a \ac{BBC} dataset show that \ac{RefNet} significantly outperforms state-of-the-art methods in terms of both automatic and, especially, human evaluations.

 Our contributions are as follows:
 \begin{itemize}[nosep]
 \item We propose a novel architecture, \ac{RefNet}, for \acp{BBC} by combing the advantages of extraction-based and generation-based methods. \ac{RefNet} can generate more informative and appropriate responses while retaining fluency. 
 \item We devise a decoding switcher and a hybrid decoder to adaptively coordinate between \emph{reference decoding} and \emph{generation decoding}.
 \item Experiments show that \ac{RefNet} outperforms state-of-the-art models by a large margin in terms of both automatic and human evaluations.
 \end{itemize}

\section{Related work}
We survey two types of related work on \acp{BBC}: generation-based and extraction-based methods.

\begin{figure*}[ht]
 \centering
 \includegraphics[width=2\columnwidth]{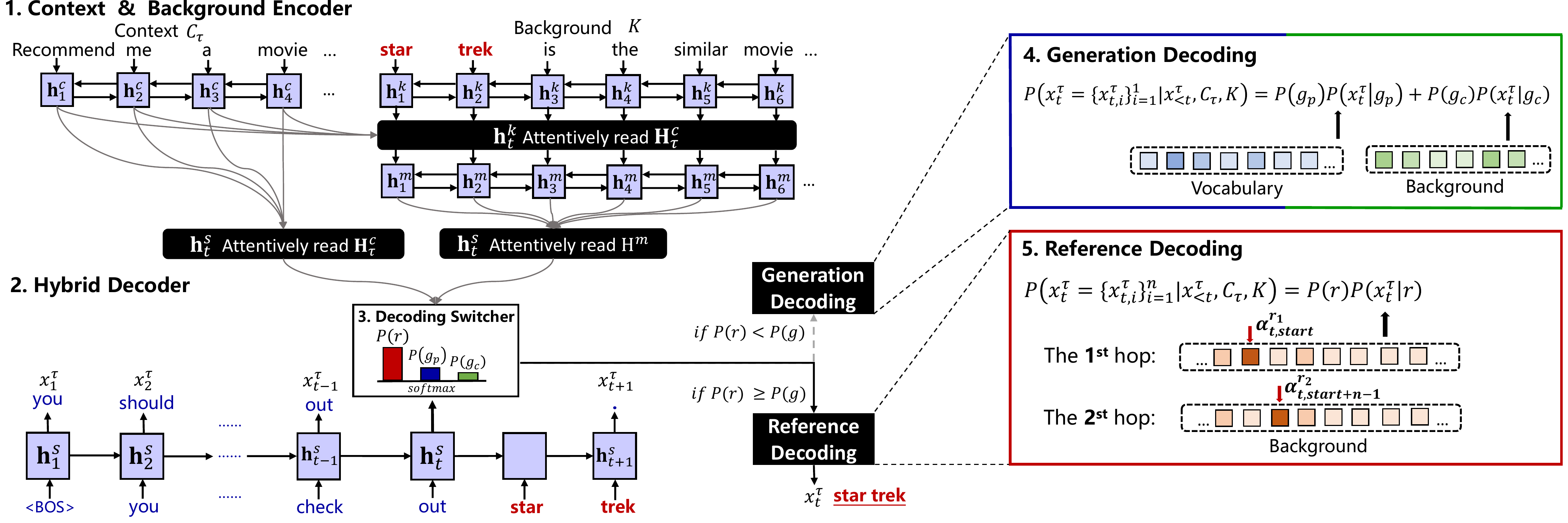}
 \caption{Overview of \ac{RefNet}.}
 \label{figure2}
\end{figure*}

\subsection{Generation-based methods}
Most effective generation-based models are based on sequence-to-sequence modeling~\citep{sutskever2014sequence} and an attention mechanism~\citep{attention2015}.
The proposed methods have achieved promising results on different conversational tasks \citep{serban2016building}.
However, response informativeness is still a urgently need to be addressed challenge; these approaches prefer generating generic responses such as "I don’t know" and "thank you", which make conversations dull~\citep{gao2019neural}.
Various methods have been proposed to improve response informativeness, such as adjusting objective functions \citep{li2016diversity,jiang-2019-improving}, incorporating latent topic information \cite{xing2017topic}, leveraging outside knowledge bases~\citep{liu2018knowledge,zhou2018commonsense} and knowledge representation~\cite{ghazvininejad2018knowledge,parthasarathi2018extending,Lian2019Learning}, etc.

Recently, \acfp{BBC} have been proposed for generating more informative responses by exploring related background information \cite{zhou2018dataset,dinan2018wizard}.
\citet{moghe2018towards} build a dataset for \ac{BBC} and conduct experiments with state-of-the-art generation-based methods.
They show that generation-based methods can generate fluent, natural responses, but have difficulty in locating the right background information.
Therefore, most recent studies try to address this issue~\citep{zekangli2018incremental,qin-etal-2019-conversing}.
\citet{zhang2019improving} introduce a pre-selection process that uses dynamic bi-directional attention to improve background information selection. 
\citet{liu2019knowledge} propose an augmented knowledge graph based chatting model via transforming background information into knowledge graph.
However, generation-based models still cannot solve inherent problems effectively, such as tending to break a complete semantic unit and generate shorter responses.

\subsection{Extraction-based methods}
Extraction-based methods have originally been proposed for \ac{RC} tasks \cite{rajpurkar2016squad}, where each question can be answered by a right span in a given passage.
\citet{wang2017machine} combine match-LSTM and a pointer network~\cite{vinyals2015pointer} to predict the boundary of the answer.
\citet{seo2016bidirectional} propose BiDAF, which uses a variant co-attention architecture \cite{Coattention} to enhance the extraction result.  
\citet{wang2017gated} propose R-Net, which introduces a self-matching mechanism.
\citet{yu2018qanet} propose QANet, which devises an encoder consisting exclusively of convolution and self-attention.
For \acp{BBC}, \citet{moghe2018towards} show that extraction-based methods are better at locating the right background information than generation-based methods. 
However, current extraction-based methods are specifically designed for \ac{RC} tasks.
They are not suitable for \acp{BBC} for two reasons:
First, \acp{BBC} usually do not have standard factoid questions like those in \ac{RC} tasks.
Second, \acp{BBC} require that the responses are fluent and conversational, which cannot be met by rigid extraction; see Fig.~\ref{f1}.

Unlike the work summarized above, we propose \ac{RefNet} to combine the advantages of generation-based and extraction-based methods while avoiding their shortcomings.
The main challenge that \ac{RefNet} addresses is how to design an effective neural architecture that is able to refer to the right background information at the right time in the right place of a conversation while minimizing the influence on response fluency.

\section{Reference-aware Network}

Given a background in the form of free text $K=(k_{1}$, $k_2$, \ldots, $k_t$, \ldots, $k_{L_{K}})$ with $L_{K}$ tokens and a current conversational context $C_{\tau}=(\ldots$, $X_{\tau-3}$, $X_{\tau-2}$, $X_{\tau-1})$, the task of \ac{BBC} is to generate a response $X_{\tau}$ at $\tau$.
Each $X_{\tau}$ contains a sequence of $L_{X_\tau}$ units, i.e., $X_{\tau}=(x^\tau_{1},x^\tau_{2},\ldots,x^\tau_{t},\ldots,x^\tau_{{L_{X_\tau}}})$, where $x^\tau_{t}$, the unit at timestamp $t$, could be a token $\{x^\tau_{t,i}\}^1_{i=1}$ or a semantic unit $\{x^\tau_{t,i}\}^n_{i=1}$ containing $n$ tokens.

\ac{RefNet} consists of four modules: background encoder, context encoder, decoding switcher, and hybrid decoder; see Fig.~\ref{figure2}.
Background and context encoders encode the given background $K$ and context $C_{\tau}$ into latent representations $\mathbf{H}^k$ and $\mathbf{H}^c_\tau$, respectively.
$\mathbf{H}^k$ and $\mathbf{H}^c_\tau$ go through a matching layer to get a context-aware background representation $\mathbf{H}^m$.
At each decoding step, the decoding switcher predicts the probabilities of executing the \emph{reference decoding} or \emph{generation decoding}.
The hybrid decoder takes $\mathbf{H}^c_\tau$, $\mathbf{H}^m$ and the embedding of the previous token as input and computes the probability of selecting a semantic unit from the background (\emph{reference decoding}) or generating a token (\emph{generation decoding}) based on the decision made by the decoding switcher.
Next, we introduce the separate modules.

\subsection{Background and context encoders}
We use a bi-directional RNN \citep{schuster1997bidirectional} with GRU \citep{cho2014learning} to convert the context and background sequences into two hidden state sequences $\mathbf{H}^c_\tau=(\mathbf{h}_1^c$, $\mathbf{h}_2^c$, \ldots, $\mathbf{h}_{L_{C_\tau}}^c)$ and $\mathbf{H}^k=(\mathbf{h}_1^k$, $\mathbf{h}_2^k$, \ldots, $\mathbf{h}_{L_K}^k)$:
\begin{equation}
\label{encoder}
\begin{split}
\mathbf{h}_t^c={}&\mathbf{BiGRU}_c(\mathbf{h}_{t-1}^c, \mathbf{e}({x_t})),\\ \mathbf{h}_t^k={}&\mathbf{BiGRU}_k(\mathbf{h}_{t-1}^k, \mathbf{e}({k_t})),
\end{split}
\end{equation}
where $\mathbf{h}_t^c$ or $\mathbf{h}_t^k$ correspond to a token in the context or background, respectively, and $\mathbf{e}(x_t)$ and $\mathbf{e}(k_t)$ are the embedding vectors, respectively.
We concatenate the responses in the context, $L_{C_\tau}$ is the number of all tokens in the context, and we do not consider the segmentation of semantic units during encoding, i.e., each $x^\tau_{t}$ is a token $\{x^\tau_{t,i}\}^1_{i=1}$. 

Further, we use a matching layer \cite{wang2017machine,wang2017gated} to get the context-aware background representation $\mathbf{H}^m=(\mathbf{h}_1^m$, $\mathbf{h}_2^m$, \ldots, $\mathbf{h}_{L_K}^m)$:
\begin{equation}
\label{context-aware encoder}
\begin{split}
\mathbf{h}_t^m={}&\mathbf{BiGRU}_m(\mathbf{h}_{t-1}^m,[\mathbf{h}_t^k;\mathbf{c}_{t}^{kc}]),
\end{split}
\end{equation}
where $\mathbf{c}_{t}^{kc}$ is calculated using an attention mechanism \citep{attention2015} with $\mathbf{h}_t^k$ attentively reading $\mathbf{H}^c_\tau$: 
\begin{equation}
\label{ba}
\begin{split}
s^{kc}_{t,j}={}&\mathbf{v}_{kc}^\mathrm{T}\tanh(\mathbf{W}_{kc}\mathbf{h}_j^c +\mathbf{U}_{kc}\mathbf{h}^{k}_{t} + \mathbf{b}_{kc}), \\
\alpha^{kc}_{t,i}={}&\frac{\exp(s^{kc}_{t,i})}{\sum_{j=1}^{L_{C_\tau}} \exp(s^{kc}_{t,j})},\ 
\mathbf{c}_{t}^{kc}={}\sum_{i=1}^{L_{C_\tau}}\alpha^{kc}_{t,i}\mathbf{h}_i^c,\\
\end{split}
\end{equation}
where $\mathbf{W}_{kc}$, $\mathbf{U}_{kc}$, $\mathbf{v}_{kc}$ and $\mathbf{b}_{kc}$ are parameters.

\subsection{Hybrid decoder}
During training, we know that the next $x^\tau_{t}$ to be generated is a token $\{x^\tau_{t,i}\}^1_{i=1}$ or a semantic unit $\{x^\tau_{t,i}\}^n_{i=1}$.
If $x^\tau_{t}=\{x^\tau_{t,i}\}^n_{i=1}$, then $x^\tau_{t}$ is generated in \emph{reference decoding} mode with the probability modeled as follows:
\begin{equation}
\label{r_dec}
P(x^\tau_{t}|x^\tau_{<t}, C_\tau, K)=P(r)P(x^\tau_{t}\mid r),
\end{equation}
where $P(r)$ is the \emph{reference decoding} probability (see \S\ref{switcher}); $P(x^\tau_{t}\mid r)$ is the probability of generating $x^\tau_{t}$ under the \emph{reference decoding} $r$. 
If $x^\tau_{t}=\{x^\tau_{t,i}\}^1_{i=1}$, then $x^\tau_{t}$ is generated in \emph{generation decoding} mode with the probability modeled as:
\begin{equation}
\label{g_dec}
\begin{split}
&P(x^\tau_{t}| x^\tau_{<t}, C_\tau, K)=\\
&\quad P(g_p)P(x^\tau_{t}\mid g_p)+P(g_c)P(x^\tau_{t}\mid g_c),
\end{split}
\end{equation}
where $P(g)=P(g_p)+P(g_c)$ is the \emph{generation decoding} probability; $P(g_p)$ is the \emph{predicting generation decoding} probability (see \S\ref{switcher}) and $P(g_c)$ is the \emph{copying generation decoding} probability (see \S\ref{switcher}).
$P(x^\tau_{t}\mid g_p)$ and $P(x^\tau_{t}\mid g_c)$ are the probabilities of generating $x^\tau_{t}$ under $g_p$ and $g_c$, respectively.

\subsubsection{Reference decoding.}
\label{rd}
Within \emph{reference decoding}, the probability of generating the semantic unit $\{x^\tau_{t,i}\}^n_{i=1}$ is evaluated as follows:
\begin{equation}
P(x^\tau_{t}=\{x^\tau_{t,i}\}^n_{i=1}|r)= \alpha^{r_1}_{t,start}\alpha^{r_2}_{t,start+n-1},
\end{equation}
where $\alpha^{r_1}_{t,start}$ and $\alpha^{r_2}_{start+n-1}$ are the probabilities of the start and end tokens of $\{x^\tau_{t,i}\}^n_{i=1}$ (from the background), respectively, which are estimated by two-hop pointers with respect to the context-aware background hidden state sequence $\mathbf{H}^m$.
The $\alpha^{r_1}_{t,start}$ is calculated by the first hop pointer, as shown in Eq.~\ref{ra1}:
\begin{equation}
\label{ra1}
\begin{split}
\mathbf{o}^{1}_t={}&\mathbf{W}_{o_1}[\mathbf{h}^{s}_{t};\mathbf{c}_{t}^{sc};\mathbf{c}_{t}^{sm}]+\mathbf{b}_{o_1}, \\
s^{r_1}_{t,j}{}={}&\mathbf{v}_{r}^\mathrm{T}\tanh(\mathbf{W}_r\mathbf{h}_j^{m} +\mathbf{U}_r\mathbf{o}^{1}_t + \mathbf{b}_r), \\
\alpha^{r_1}_{t,start}={}& \frac{\exp(s^{r_1}_{t,start})}{\sum_{j=1}^{L_{K}} \exp(s^{r_1}_{t,j})}, \\
\end{split}
\end{equation}
where $\mathbf{W}_{o_1}$, $\mathbf{W}_r$, $\mathbf{U}_r$, $\mathbf{v}_{r}$, $\mathbf{b}_{o_1}$ and $\mathbf{b}_r$ are parameters.
$\mathbf{h}^{s}_{t}$ is the decoding hidden state vector, the updating scheme of which will be detailed in \S\ref{su}.
$\mathbf{c}_{t}^{sc}$ and $\mathbf{c}_{t}^{sm}$ are calculated in a similar way like Eq.~\ref{ba} with $\mathbf{h}^{s}_{t}$ attentively reading $\mathbf{H}^c_\tau$ and $\mathbf{H}^m$, respectively.
The $\alpha^{r_2}_{t,start+n-1}$ is calculated by the second hop pointer, as shown in Eq.~\ref{ra2}:
\begin{equation}
\label{ra2}
\begin{split}
\mathbf{c}_{t}^{r}={}&\sum_{i=1}^{L_{K}}\alpha^{r_1}_{t,i}\mathbf{h}_i^m,\  \mathbf{o}^{2}_t{}=\mathbf{W}_{o_2}[\mathbf{o}^{1}_t;\mathbf{c}_{t}^{r}]+\mathbf{b}_{o_2},\\
s^{r_2}_{t,j}={}&\mathbf{v}_{r}^\mathrm{T}\tanh(\mathbf{W}_r\mathbf{h}_j^{m} +\mathbf{U}_r\mathbf{o}^{2}_t + \mathbf{b}_r), \\
\mbox{}\hspace*{-2mm}\alpha^{r_2}_{t,start+n-1}={}&\frac{\exp(s^{r_2}_{t,start+n-1})}{\sum_{j=1}^{L_{K}} \exp(s^{r_2}_{t,j})},\\
\end{split}
\end{equation}
where $\mathbf{W}_{o_2}$ and $\mathbf{b}_{o_2}$ are parameters.
\emph{Reference decoding} adopts soft pointers $\alpha^{r_1}_{t,start}$ and $\alpha^{r_2}_{start+n-1}$ to select semantic units, so it will not influence the automatic differentiation during training.

\subsubsection{Generation decoding.}
\label{gd}
Within \emph{predicting generation decoding}, the probability of predicting the token $x^\tau_{t}$ from the vocabulary is estimated as follows:
\begin{equation} 
\begin{split}
\mbox{}\hspace*{-2mm}P(x^\tau_{t}=\{x^\tau_{t,i}\}^1_{i=1} {\mid} g_p)\!=\!\mathrm{softmax}(\mathbf{W}_{g_p}\mathbf{o}^{1}_t+\mathbf{b}_{g_p}),\,
\end{split}
\end{equation}
where $\mathbf{W}_{g_p}$ and $\mathbf{b}_{g_p}$ are parameters and the vector $\mathbf{o}^{1}_t$ is the same one as in Eq.~\ref{ra1}. 

Within \emph{copying generation decoding}, the probability of copying the token $x^\tau_{t}$ from the background is estimated as follows:
\begin{equation}
\begin{split}
P(x^\tau_{t}=\{x^\tau_{t,i}\}^1_{i=1}\mid g_c)=\sum_{i:k_i=x^\tau_{t}}\alpha^{sm}_{t,i}, 
\end{split}
\end{equation}
where $\alpha^{sm}_{t,i}$ is the attention probability distribution on $\mathbf{H}^m$ produced by the same attention process with $\mathbf{c}_{t}^{sm}$ in Eq.~\ref{ra1}.
        
\subsection{Decoding switcher}
\label{switcher}
The decoding switching probabilities $P(r)$, $P(g_p)$ and $P(g_c)$ are estimated as follows:
\begin{equation}
\label{fusion vector}
\begin{split}
[P(r),P(g_p),P(g_c)]=\mathrm{softmax}(\mathbf{f}_t),
\end{split}
\end{equation}
where $\mathbf{f}_t$ is a fusion vector, which is computed through a linear transformation in Eq.~\ref{fv}:
\begin{equation}
\label{fv}
\begin{split}
\mathbf{f}_t=\mathbf{W}_f[\mathbf{h}^{s}_{t};\mathbf{c}_{t}^{sc};\mathbf{c}_{t}^{sm}]+\mathbf{b}_f,
\end{split}
\end{equation}
where $\mathbf{W}_f$ and $\mathbf{b}_f$ are parameters.
$\mathbf{h}^{s}_{t}$ is decoding states (see \S\ref{su}).

During testing, at each decoding step, we first compute $P(r)$ and $P(g)=P(g_p)+P(g_c)$.
If $P(r) \geq P(g)$, we use Eq.~\ref{r_dec} to generate a semantic unit, otherwise we use Eq.~\ref{g_dec} to generate a token.

\subsection{State updating}
\label{su}
The decoding state updating depends on whether the generated unit is a token or semantic unit. 
If $x^\tau_{t-1}$ is a token, then $\mathbf{h}^{s}_{t}=$
\begin{equation}
\label{tsu}
\begin{split}
&\mbox{}\hspace*{-1mm}\mathbf{GRU}(\mathbf{h}^{s}_{t-1},[\mathbf{e}(x^\tau_{t-1});\mathbf{c}_{t-1}^{sc};\mathbf{c}_{t-1}^{sm}]).\hspace*{-2mm}\mbox{}
\end{split}
\end{equation}
If $x^\tau_{t-1}$ is a span containing $n$ tokens, Eq.~\ref{tsu} will update $n$ times with one token as the input, and the last state will encode the full semantics of a span; see $\mathbf{h}^{s}_{t}$ to $\mathbf{h}^{s}_{t+1}$ in Fig.~\ref{figure2}. 

The decoding states are initialized using a linear layer with the last state of $\mathbf{H}^m$ and $\mathbf{H}^c_\tau$ as input:  
\begin{equation}
\label{states initialize}
\begin{split}
\mathbf{h}^{s}_{0}&=\text{ReLU}(\mathbf{W}_{hs}[\mathbf{h}_{L_K}^m;\mathbf{h}_{L_{C_\tau}}^c]+\mathbf{b}_{hs}),
\end{split}
\end{equation}
where $\mathbf{W}_{hs}$ and $\mathbf{b}_{hs}$ are parameters.
ReLU is the ReLU activation function.

\subsection{Training}
Our goal is to maximize the prediction probability of the target response given the context and background.
We have three objectives, namely generation loss, reference loss and switcher loss.

The \emph{generation loss} is defined as $\mathcal{L}_{g}(\theta)=$
\begin{equation}
\label{gen_loss}
\begin{split}
-\frac{1}{M}\sum_{\tau=1}^{M}\sum_{t=1}^{{L_{{X}_\tau}}}\log[P(x^\tau_{t}\mid x^\tau_{<t},C_{\tau},K)],
\end{split}
\end{equation}
where $\theta$ are all the parameters of \ac{RefNet}.
$M$ is the number of all training samples given a background $K$.
In $\mathcal{L}_{g}(\theta)$, each $x^\tau_{t}$ is a token $\{x^\tau_{t,i}\}^1_{i=1}$.

The \emph{reference loss} is defined as $\mathcal{L}_{r}(\theta)=$
\begin{equation}
\label{ref_loss}
\begin{split}
-\frac{1}{M}\sum_{\tau=1}^{M}\sum_{t=1}^{{L_{{X}_\tau}}}&I(x^\tau_{t})\cdot
\log[P(x^\tau_{t}\mid x^\tau_{<t},C_{\tau},K))],\hspace*{-2mm}\mbox{}
\end{split}
\end{equation}
where $I(x^\tau_{t})$ is an indicator function that equals 1 if $x^\tau_{t}=\{x^\tau_{t,i}\}^n_{i=1}$ and 0 otherwise. 

\ac{RefNet} introduces a decoding switcher to decide between \emph{reference decoding} and \emph{generation decoding}.
To better supervise this process we define \emph{switcher loss} $\mathcal{L}_{s}(\theta)=$
\begin{equation}
\label{switcher_loss} 
\begin{split}
\mbox{}\hspace*{-2mm}
{-}\frac{1}{M}\sum_{\tau=1}^{M}\sum_{t=1}^{{L_{{X}_\tau}}} I&(x^\tau_{t}) \log[P(r)]\!+\!
(1{-}I(x^\tau_{t}))\log[P(g)],
\hspace*{-1mm}\mbox{}
\end{split}
\end{equation}
where $I(x^\tau_{t})$ is also an indicator function, which is the same as in $\mathcal{L}_{r}(\theta)$.

The \emph{final loss} is a linear combination of the three loss functions just defined:
\begin{equation}
\label{loss}
\begin{split}
\mathcal{L}(\theta)=\mathcal{L}_{g}(\theta)+\mathcal{L}_{r}(\theta)+\mathcal{L}_{s}(\theta).
\end{split}
\end{equation}
All parameters of \ac{RefNet} as well as word embeddings are learned in an end-to-end back-propagation training paradigm.

\section{Experimental Setup}

\subsection{Implementation details}
We set the word embedding size and GRU hidden state size to 128 and 256, respectively.
The vocabulary size is limited to 25,000.
For fair comparison, all models use the same embedding size, hidden state size and vocabulary size.
Following \citet{moghe2018towards}, we limit the context length of all models to 65.
We train all models for 30 epochs and test on a validation set after each epoch, and select the best model based on the validation results according to BLEU metric.
We use gradient clipping with a maximum gradient norm of 2.
We use the Adam optimizer with a mini-batch size of 32.
The learning rate is 0.001.
The code is available online.\footnote{https://github.com/ChuanMeng/RefNet}

\subsection{Dataset}
Recently, some datasets for \acp{BBC} have been released ~\citep{zhou2018dataset,dinan2018wizard}.
We choose the Holl-E dataset released by \citet{moghe2018towards} because it contains boundary annotations of the background information used for each response.
We did not use the other datasets because they do not have such annotations for training \ac{RefNet}.
Holl-E is built for movie chats in which each response is explicitly generated by copying and/or modifying sentences from the background.
The background consists of plots, comments and reviews about movies collected from different websites.
We use the mixed-short background which is truncated to 256 words, because it is more challenging according to \citet{moghe2018towards}.
We follow the original data split for training, validation and test.
There are also two versions of the test set: one with single golden reference (SR) and the other with multiple golden references (MR);
see \citep{moghe2018towards}.

\subsection{Baselines}

We compare with all methods we can get on this task.

\begin{itemize}[nosep]
\item Extraction-based methods\footnote{For fair comparison, different from \citet{moghe2018towards}, we do not use pre-trained GloVe~\citep{pennington2014glove} such that all models randomly initialize the word embedding with the same vocabulary size.}:
\begin{enumerate*}[label=(\roman*)]
\item \textbf{BiDAF} extracts a span from background as response and uses a co-attention architecture to improve the span finding accuracy~\citep{seo2016bidirectional}.
\item \textbf{R-Net} proposes gated attention-based recurrent networks and a self-matching attention mechanism to encode background~\citep{wang2017gated}.
\item \textbf{QANet} uses an encoder consisting exclusively of convolution and self-attention to capture local and global interactions in background~\citep{yu2018qanet}.
\end{enumerate*}


\item Generation-based methods: 
\begin{enumerate*}[label=(\roman*)] 
\item \textbf{S2S} maps the context to the response with an encoder-decoder framework \citep{sutskever2014sequence}.
\item \textbf{HRED} encodes the context of the conversation with two hierarchical levels \citep{serban2016building}. 
 S2S and HRED do not use any background information.
\item \textbf{S2SA} adds an attention mechanism to the original S2S model to attend to the relevant background information \citep{attention2015}.
\item \textbf{GTTP} leverages background information with a copying mechanism to copy a token from the background at the appropriate decoding step \citep{see2017get}.
\item \textbf{CaKe} is a improved version of GTTP, which introduces a pre-selection process that uses dynamic bi-directional attention to improve knowledge selection from background \citep{zhang2019improving}.
\item \textbf{AKGCM} first transforms background information into knowledge graph, and uses a policy network to select knowledge with an additional GTTP to generate responses \citep{liu2019knowledge}.
\end{enumerate*}
\end{itemize}

\subsection{Evaluation metrics}
Following the work of \citet{moghe2018towards}, we use BLEU-4, ROUGE-1, ROUGE-2 and ROUGE-L as automatic evaluation metrics.
We also report the average length of responses outputted by each model.
For extraction-based methods and \ac{RefNet}, we further report F1 \citep{seo2016bidirectional}, which only evaluates the extracted spans not the whole responses.
We also randomly sample 500 test samples to conduct human evaluations using Amazon Mechanical Turk.
For each sample, we ask 3 workers to annotate whether the response is good in terms of four aspects: (1)~\emph{Naturalness} (\textbf{N}), i.e., whether the responses are conversational, natural and fluent; (2)~\emph{Informativeness} (\textbf{I}), i.e., whether the responses use some background information; (3)~\emph{Appropriateness} (\textbf{A}), i.e., whether the responses are appropriate/relevant to the given context; and (4)~\emph{Humanness} (\textbf{H}), i.e., whether the responses look like they are written by a human.

\begin{table*}[th]
\centering
\caption{
Automatic evaluation results.
\textbf{Bold face} indicates leading results.
Significant improvements over the best baseline results are marked with $^\ast$ (t-test, $p < 0.05$).
SR and MR refer to test sets with single and multiple references.
The results of AKGCM are taken from the paper because the authors have not released their code and processed knowledge graph.
Note that AKGCM uses GloVe and BERT to improve performance but none of other models do.
}
\label{result1}
\resizebox{1.98\columnwidth}{!}{
\begin{tabular}{l cc cc cc cc cc cc}
\toprule
\multirow{2}{*}{\textbf{Methods}} & \multicolumn{2}{c}{\textbf{F1}} & \multicolumn{2}{c}{\textbf{BLEU}}                                & \multicolumn{2}{c}{\textbf{ROUGE-1}}                             & \multicolumn{2}{c}{\textbf{ROUGE-2}}                             & \multicolumn{2}{c}{\textbf{ROUGE-L}} & \multirow{2}{*}{\textbf{Average length}}                             \\ 
\cmidrule(lr){2-3} \cmidrule(lr){4-5} \cmidrule(lr){6-7} \cmidrule(lr){8-9} \cmidrule(lr){10-11}  
                                  & \textbf{SR}     & \textbf{MR}    & \textbf{SR}                     & \textbf{MR}                     & \textbf{SR}                     & \textbf{MR}                     & \textbf{SR}                     & \textbf{MR}                     & \textbf{SR}                     & \textbf{MR}                     \\ 
\midrule
\multicolumn{11}{c}{\textbf{no background}}                                                                                                                                                                                                                                                                                                        \\ \midrule
S2S                               & -               & -              & 5.26                            & 7.11                            & 27.15                           & 30.91                           & 9.56                          & 11.85                           & 21.48                           & 24.81               & 16.08            \\ 
HRED                              & -               & -              & 5.23                            & 5.38                            & 24.55                           & 25.38                           & 7.61                            & 8.35                            & 18.87                           & 19.67   & 16.22                         \\ \midrule

\multicolumn{11}{c}{\textbf{mixed-short background~(256 words)}}                        \\ \midrule
BiDAF   & 40.38  & 45.86  & 27.44 & 33.40 & 38.79 & 43.93 & 32.91   & \textbf{39.50}     & 35.09  & 40.12 & 25.40       \\   
R-Net   & 40.92 & 46.84  & 27.54 & 33.18 & 39.78 & 44.30 & 32.34 & 37.65 & 35.63 & 40.49 & 23.08\\
QANet  & 41.65 & 47.32 & 28.21  & 33.91  & 40.66 & 44.82   & \textbf{33.62} & 39.04 & 35.29 & 41.02 & 23.21   \\
\midrule
S2SA                              & -               & -              & 11.71                           & 12.76                                & 26.36        & 30.76          & 13.36         & 16.69       & 21.96                           & 25.99                        & 16.94    \\ 
GTTP                              & -               & -              & 13.65                           & 19.49                           & 30.77                           & 36.06                           & 18.72                           & 23.70                           & 25.67                           & 30.69     & 14.31        \\ 
CaKe  & - & - & 26.03 & 29.18 & 40.21 & 44.12 & 29.03 & 34.00 & 35.01 & 39.03   & 20.06 \\
AKGCM
& - & -  & \textbf{30.84} & -   & - & - & 29.29 & -  & 34.72 & - & -\\
\midrule
RefNet    & \textbf{41.86}\rlap{$^\ast$} & \textbf{48.46}\rlap{$^\ast$}   & 30.33 & \textbf{33.97} & \textbf{42.11}\rlap{$^\ast$} &\textbf{47.35}\rlap{$^\ast$} & 31.35 & 36.53 & \textbf{36.70}\rlap{$^\ast$} & \textbf{41.88}\rlap{$^\ast$} & 23.51\\ \midrule
\end{tabular}
}
\end{table*}

\section{Results}
\subsection{Automatic evaluation}
We list the results of all methods on mixed-short background setting in Table~\ref{result1}.

First, \ac{RefNet} significantly outperforms all generation-based methods on all metrics, except in the BLEU score compared to AKGCM.
Especially, \ac{RefNet} outperforms the recent and strong baseline CaKe by around 2\%-4\% (significantly).
The improvements show that \ac{RefNet} is much better at leveraging and locating the right background information to improve responses than these generation-based methods.
We believe \ac{RefNet} benefits from \emph{reference decoding} to tend to produce more complete semantic units, alleviating the inherent problems that pure generation-based method faced.

Second, \ac{RefNet} outperforms extraction-based methods in most cases, including the strong baseline QANet.
We think the reason is that extraction-based methods can only rigidly extracts the relevant spans from the background, which does not consider the conversational characteristics of responses.
Differently, \ac{RefNet} also benefits from the \emph{generation decoding} to generate natural conversational words in responses, which makes up the shortcoming of only extraction.
\ac{RefNet} is comparable in average length with extraction-based methods, which demonstrates that \ac{RefNet} retains the advantages of extraction-based methods.

Third, the performance of these three extraction-based methods are comparable. 
However, their performances differ greatly between each other on the \ac{RC} task dataset SQuAD \cite{rajpurkar2016squad}, e.g. QANet outperforms BiDAF by around 7\% on F1 score.
Even with a stronger extraction-based model, we will arrive at a similar conclusion that they cannot generate natural and fluent responses due to the extraction nature.
This confirms that extraction-based methods are not suitable for this task.
Besides, we can further enhance the \emph{reference decoding} of \ac{RefNet} by incorporating various mechanisms used by extraction-based models.
But that's beyond the scope of this paper.

\subsection{Human evaluation}

\begin{table}[th]
\centering
\small
\caption{
Human evaluation results on mixed-short background version.
$\geq n$ means that at least $n$ MTurk workers think it is a good response w.r.t. \emph{Naturalness} (\textbf{N}), \emph{Informativeness} (\textbf{I}), \emph{Appropriateness} (\textbf{A}) and \emph{Humanness} (\textbf{H}).
}
\label{result2}
\begin{tabular}{lcccccc}
\toprule
\multirow{2}{*}{} & \multicolumn{2}{c}{CaKe} & \multicolumn{2}{c}{QANet} & \multicolumn{2}{c}{RefNet} \\ 
\cmidrule(lr){2-3}\cmidrule(lr){4-5}\cmidrule(lr){6-7}
                  & $\geq 1$      & $\geq 2$      & $\geq 1$       & $\geq 2$      & $\geq 1$       & $\geq 2$       \\ \midrule
(\textbf{N})      & 449            & 264       & 288        & 63      & \textbf{457}     & \textbf{299}           \\ 
(\textbf{I})      & 359            & 115             & 414      & 225   & \textbf{434}  & \textbf{247}   \\ 
(\textbf{A})      & 390             & 153             & 406     & 213 & \textbf{435}      & \textbf{240}             \\ 
(\textbf{H})      & 438         & 231            & 355          & 128      & \textbf{444}    & \textbf{242}            \\ \bottomrule
\end{tabular}
\end{table}

We also conduct a human evaluation for \ac{RefNet}, CaKe (the best generation-based baseline), and QANet (the best extraction-based baseline).
The results are shown in Table~\ref{result2}.
Generally, \ac{RefNet} achieves the best performance in terms of all metrics.
In particular, we find that \ac{RefNet} is even better than CaKe in terms of \textit{Naturalness} and \textit{Humanness}.
We believe this is because \ac{RefNet} has a good trade-off between \emph{reference decoding} and \emph{generation decoding}, where the generated conversational words and the selected semantic units are synthesized in a natural and appropriate way.
\ac{RefNet} is also much better than CaKe in terms of \textit{Appropriateness} and \textit{Informativeness}, which shows that \ac{RefNet} is better at locating the appropriate semantic units.
The reason is that with the ability to generate a full semantic unit at once, \ac{RefNet} has a global perspective to locate the appropriate semantic units, reducing the risk of breaking a complete semantic unit.
QANet achieves good evaluation scores on \textit{Informativeness} and \textit{Appropriateness} than CaKe, but gets the worst scores on \textit{Naturalness} and \textit{Humanness}.
Although QANet is relatively good at locating the relevant semantic unit, its responses lack contextual explanations, which makes workers hard to understand.
This further shows that only extracting a span from the background is far from enough for \acp{BBC}, even replacing QANet with a more stronger extraction-based one.

\section{Analysis}
\subsection{Reference vs. generation decoding}

\begin{table}[t]
\centering
\small
\caption{Analysis of reference and generation decoding on mixed-short background version.
\textbf{Bold face} indicates leading results.
Significant improvements over the best competitor are marked with $^\ast$ (t-test, $p < 0.05$).}
\label{analysis1}
\resizebox{1\columnwidth}{!}{ 
\begin{tabular}{lcccccc}
\toprule
\multirow{2}{*}{} & \multicolumn{2}{c}{Force reference} & \multicolumn{2}{c}{Force generation} & \multicolumn{2}{c}{Combination}  \\ 
\cmidrule(lr){2-3}\cmidrule(lr){4-5}\cmidrule(lr){6-7} 
                  & SR             & MR             & SR            & MR    & SR            & MR        \\ 
\midrule
BLEU              & 26.73          & 30.84          & 26.01         & 29.19   & \textbf{30.33}\rlap{$^\ast$}         & \textbf{33.97}\rlap{$^\ast$}       \\ 
ROUGE-1           & 38.03          & 43.76          & 39.86         & 45.53    & \textbf{42.11}\rlap{$^\ast$}         & \textbf{47.35}\rlap{$^\ast$}       \\ 
ROUGE-2           & 29.06          & 34.70          & 28.34         & 34.07    & \textbf{31.35}\rlap{$^\ast$}         & \textbf{36.53}\rlap{$^\ast$}       \\ 
ROUGE-L           & 34.11          & 39.67          & 35.03         & 40.63     & \textbf{36.70}\rlap{$^\ast$}         & \textbf{41.88}\rlap{$^\ast$}        \\ 
\bottomrule
\end{tabular}
}
\end{table}

\begin{table*}[t]
\centering
\small
\caption{
Case study.
\textbf{Bold face} indicates the true span in the current turn.
}
\label{analysis3}
\resizebox{2\columnwidth}{!}{
\begin{tabular}{lm{7.7cm}m{7.7cm}}
\toprule
       & \normalsize \textbf{Example 1}   & \normalsize\textbf{Example 2} \\ 
       \midrule
       & \textbf{Background}: ... but if you like ben stiller , go see " meet the fockers " . dustin 's antics will favorite character was jack ( the older one ) , because he was so serious but always plotting and putting up a front . i think it was \textbf{\$ 279,167,575} awards ascap film and television music awards 2005 top box office films mtv ... & \textbf{Background}: ...being captured by boris and onatopp . \textbf{bond arrives in st . petersburg and meets his cia contact , jack wade ( joe don baker ) .} wade agrees to take bond to the hideout of a russian gangster , valentin zukovsky ( robbie coltrane ) , whom bond had shot in the leg and given a permanent limp years before ...\\ 
       \midrule
       & \multicolumn{1}{l}{\begin{tabular}[l]{@{}p{6.8cm}@{}}\textbf{Human1}: that name is so ridiculous but funny . \\ \textbf{Human2}: first off , the writers did not miss a single opportunity to play off of the name " focker " . \\ \textbf{Human1}: yeah , i heard it was a pretty successful movie overall .\end{tabular}}  & \multicolumn{1}{l}{\begin{tabular}[l]{@{}p{6.8cm}@{}}\textbf{Human1}: that was a good seen . \\ \textbf{Human2}: what did you like about the movie ?\\ \textbf{Human1}: i liked his friend , jack wade .\\  \end{tabular}}\\ 
       \midrule
CaKe & i agree , ben stiller , go see `` meet the fockers '' .  & let them pout and go back to macgyver reruns . \\
	\midrule
QANet & \textbf{\$279,167,575}                   & \textbf{bond arrives in st. petersburg and meets his cia contact, } \textbf{jack wade (joe don baker).}            \\ 
	\midrule
RefNet & it made \textbf{\$ 279,167,575} at the box office .  & i loved the part where \textbf{bond arrives in st . petersburg and } \textbf{meets his cia contact , jack wade ( joe don baker ) .} \\
	\bottomrule
\end{tabular}
}
\end{table*}

To analyze the effectiveness of reference and generation decoding, we compare the results of \ac{RefNet} with only reference decoding (\emph{force reference}) and with only generation decoding (\emph{force generation}) in Table~\ref{analysis1}.
Note that \emph{force generation} is better than GTTP because there are two differences\footnote{We use the code released by \citet{moghe2018towards} https://github.com/nikitacs16/Holl-E}.
First, we use a matching layer to get the context-aware background representation in Eq.~\ref{context-aware encoder}, while GTTP only uses basic background representations without such a matching operation. 
Second, we use the hidden states of the background and context to jointly initialize the decoding states in Equation Eq.~\ref{states initialize}, while GTTP only uses the single representation of background to initialize it. 
We can see that force reference and force generation are comparable if working alone.
The contributions of reference and generation decoding are complementary as the combination brings further improvements on all metrics, demonstrating the need for both.

\subsection{Switcher loss} 

\begin{table}[t]
\centering
\small
\caption{
Analysis of switcher loss on mixed-short background version.
\textbf{Bold face} indicates leading results.
Significant improvements over the best competitor are marked with $^\ast$ (t-test, $p < 0.05$). 
}
\label{analysis2}
\begin{tabular}{lcccc}
\toprule
\multirow{2}{*}{} & \multicolumn{2}{c}{Without SL} & \multicolumn{2}{c}{With SL} \\ 
\cmidrule(lr){2-3}\cmidrule(lr){4-5} 
                  & SR             & MR               & SR            & MR         \\ 
	\midrule
F1                & 37.13          & 43.42          & \textbf{41.86}\rlap{$^\ast$}         & \textbf{48.46}\rlap{$^\ast$}            \\ 
BLEU              & 28.96           & 31.63         & \textbf{30.33}\rlap{$^\ast$}         & \textbf{33.97}\rlap{$^\ast$}             \\ 
ROUGE-1           & 41.27          & 46.67         & \textbf{42.11}\rlap{$^\ast$}         & \textbf{47.35}\rlap{$^\ast$}              \\ 
ROUGE-2           & 30.65           & 35.98         & \textbf{31.35}\rlap{$^\ast$}         & \textbf{36.53}            \\ 
ROUGE-L           & 36.02           & 41.86          & \textbf{36.70}\rlap{$^\ast$}         & \textbf{41.88}             \\ 

\bottomrule
\end{tabular}
\end{table}

To verify the effectiveness of the switcher loss $\mathcal{L}_{s}(\theta)$ in Eq.~\ref{switcher_loss}, we compare \ac{RefNet} with and without training switcher loss, as shown in Table~\ref{analysis2}.
We find that the overall performance increases in terms of all metrics with switcher loss, especially on F1.
It means that the switcher loss is an effective component, which better guides the model to choose between \emph{reference decoding} and \emph{generation decoding} at the right time in the right place of a conversation by additional supervision signal.
The obvious increase of F1 further shows that at the right time to cite a semantic unit may bring higher accuracy.

\subsection{Case study}

We select some examples from the test set to illustrate the performance of \ac{RefNet}, CaKe and QANet, as shown in Table~\ref{analysis3}.
One can see that \ac{RefNet} can select the right semantic unit from the background or generate fluent tokens at appropriate time and position, resulting in more informative and appropriate responses.
For instance, in Example~1, \ac{RefNet} identifies the right semantic unit ``\$279,167,575'' within the background, which is combined with ``it made'' ahead and followed by ``at the box office'' to form a more natural and conversational response.
The second example indicates that \ac{RefNet} can locate longer semantic units accurately.
In contrast, the responses by QANet lack naturality.
The responses by CaKe are relatively inconsistent and irrelevant.
In the first example, CaKe breaks the complete semantic unit ``if you like ben stiller '' and throws out the part ``if you like ''.

There are also some cases where \ac{RefNet} does not perform well.  
For example, we find that \ac{RefNet} occasionally selects short or meaningless semantic units, such as ``i'' and ``it.''
This indicates that we could further improve reference decoding by taking more factors (e.g., the length of semantic units) into consideration.

\section{Conclusion and Future Work}
In this paper, we propose \ac{RefNet} for the Background Based Conversation (\acp{BBC}) task.
\ac{RefNet} incorporates a novel \emph{reference decoding} module to generate more informative responses while retaining the naturality and fluency of responses.
Experiments show that \ac{RefNet} outperforms state-of-the-art methods by a large margin in terms of both automatic and human evaluations.

A limitation of \ac{RefNet} is that it needs boundary annotations of semantic units to enable supervised training.
In future work, we hope to design a weakly supervised or unsupervised training scheme for \ac{RefNet} in order to apply it to other datasets and tasks.
In addition, we will consider more factors (e.g., the length or frequency of semantic unit) to further improve the reference decoding module of \ac{RefNet}.

\section*{Acknowledgments}
We thank the anonymous reviewers for their helpful comments.
This work is supported by the Natural Science Foundation of China (61972234, 61902219, 61672324, 61672322), the Natural Science Foundation of Shandong province (2016ZRE27468), the Tencent AI Lab Rhino-Bird Focused Research Program (JR201932), the Fundamental Research Funds of Shandong University, Ahold Delhaize, the Association of Universities in the Netherlands (VSNU), and the Innovation Center for Artificial Intelligence (ICAI).

\bibliography{bibtex}
\bibliographystyle{aaai}

\end{document}